\pdfoutput=1

\documentclass[11pt]{article}

\usepackage{tabularx}
\usepackage{multirow}
\usepackage{arydshln}
\usepackage{microtype}
\usepackage{mwe}

\newcolumntype{Y}{>{\centering\arraybackslash}X}
\usepackage{amsmath}
\usepackage{comment}
\usepackage{booktabs}
\usepackage{amsfonts}
\usepackage{adjustbox}

\usepackage{acl}

\usepackage{times}
\usepackage{latexsym}

\usepackage[T1]{fontenc}

\usepackage[utf8]{inputenc}

\usepackage{microtype}

\usepackage{todonotes}
\makeatletter
\newcommand*\iftodonotes{\if@todonotes@disabled\expandafter\@secondoftwo\else\expandafter\@firstoftwo\fi}
\makeatother

\definecolor{edolime}{rgb}{0.9,1,0.3}

\usepackage{stmaryrd}

\title{Parameter-Efficient Neural Reranking for \\ Cross-Lingual and Multilingual Retrieval}

\author{Robert Litschko$^{\mathbf{1}}$ ~~ Ivan Vuli\'{c}$^{\mathbf{2}}$ ~~ {Goran Glava\v{s}}$^{\mathbf{1,3}}$\\
$^{\mathbf{1}}$ Data and Web Science Group, University of Mannheim, Germany \\
$^{\mathbf{2}}$ Language Technology Lab, University of Cambridge, UK \\
$^{\mathbf{3}}$ Center for AI and Data Science (CAIDAS), University of Würzburg, Germany }

\begin{document}
\maketitle
\begin{abstract}
State-of-the-art neural (re)rankers are notoriously data-hungry which -- given the lack of large-scale training data in languages other than English -- makes them rarely used in multilingual and cross-lingual retrieval settings. Current approaches therefore commonly transfer rankers trained on English data to other languages and cross-lingual setups by means of multilingual encoders: they fine-tune \textit{all} parameters of pretrained massively multilingual Transformers (MMTs, e.g., multilingual BERT) on English relevance judgments, and then deploy them in the target language(s).    
In this work, we show that two \textit{parameter-efficient} approaches to cross-lingual transfer, namely Sparse Fine-Tuning Masks (SFTMs) and Adapters, allow for a \textit{more lightweight} and \textit{more effective} zero-shot transfer to multilingual and cross-lingual retrieval tasks. We first train language adapters (or SFTMs) via Masked Language Modelling and then train retrieval (i.e., reranking) adapters (SFTMs) on top, while keeping all other parameters fixed. At inference, this modular design allows us to compose the ranker by applying the (re)ranking adapter (or SFTM) trained with source language data together with the language adapter (or SFTM) of a target language. 
We carry out a large scale evaluation on the CLEF-2003 and HC4 benchmarks and additionally, as another contribution, extend the former with queries in three new languages: Kyrgyz, Uyghur and Turkish. The proposed parameter-efficient methods 
outperform standard zero-shot transfer with full MMT fine-tuning, while being more modular and reducing training times. The gains are particularly pronounced for low-resource languages, where our approaches also substantially outperform the competitive machine translation-based rankers.
\end{abstract}

\section{Introduction}
\label{s:intro}


In recent years, neural rankers \cite{nogueira2019multi,macavaney2019cedr,khattab2020colbert}, trained on large-scale datasets \cite{bajaj2016ms,dietz2017trec,craswell2021ms}, have substantially pushed the performance on various retrieval benchmarks. Since such models are generally too computationally involved (i.e., too slow) for ad-hoc retrieval on large document collections, they are commonly leveraged as rerankers, i.e., they rerank the output of some fast model (e.g., BM25) that produces the initial ranking. Large-scale datasets for training neural rerankers, however, exist only in English, which impedes their adoption in retrieval scenarios that involve other languages: (a) monolingual retrieval in other languages and (b) cross-lingual information retrieval (CLIR) in which,  
for a given query in one language, one needs to determine relevance of documents written in one or more other languages. 

While CLIR is often instantiated in the form of standalone tasks (e.g.,  
to allow users from different countries to search over the aggregated global collection of COVID-19 news and findings in their native language \cite{casacuberta2021covid}), it also supports a range of IR-backed NLP tasks such as cross-lingual question answering \cite{asai-etal-2021-xor}, entity linking \cite{quian2021xlingentitylinking}, and cross-lingual summarization \cite{zhu2019ncls,vitiugin2022cross}.
A truly multilingual search engine requires reliable estimation of both monolingual (for a wide range of languages) as well as cross-lingual query-document relevance, which both crucially rely on the alignment of text representations across different languages \cite{nie:2010book}. The lack of large-scale retrieval datasets in languages other than English means that monolingual reranking for those languages has to be achieved by means of cross-lingual transfer of a reranking model trained on English relevance judgments. 

Pretrained massively multilingual Transformers (MMTs) like multilingual BERT (mBERT) \cite{devlin-etal-2019-bert} or XLM-R \cite{conneau2020unsupervised} have been leveraged to this effect, but were shown to require substantial task-specific (i.e., ranking-oriented) fine-tuning for reliable prediction of semantic similarity and relevance scores \cite{reimers2020distil,litschko2021encoderclir}.
MMTs offer zero-shot cross-lingual transfer of neural (re)ranking models out of the box -- an MMT is fine-tuned on English relevance judgments and then employed in (monolingual or cross-lingual) retrieval tasks that involve other languages. Conceptually, via such transfer, no fine-tuning data (i.e., relevance judgments) is required for the target language(s). 

This procedure, in principle, enables downstream zero-shot transfer to any language seen by the MMT in pretraining (e.g., for mBERT, 104 languages). However, in language understanding tasks \cite{hu2020xtreme}, massive performance drops have been observed when transferring between distant languages, and especially in transfer to low-resource languages, underrepresented in MMT pretraining \cite{lauscher2020zero}. Our results (\S\ref{s:results}) confirm these findings for ad-hoc IR.   
This is the consequence of the effect known as the \textit{curse of multilinguality} \cite{conneau2020unsupervised}: sharing MMT parameters (i.e., its fixed parameter budget/capacity) across more and more languages makes text representations for individual languages less accurate. This effect is especially detrimental to low-resource languages, those least represented in multilingual pretraining corpora. What is more, large-scale full fine-tuning on the source language data (e.g., English) is likely to lead to catastrophic forgetting and interference effects \cite{mccloskey1989catastrophic,mirzadeh2020understanding} that further bias the multilingual representation space towards the source language, at the expense of representation quality for low-resource languages.  
%
Besides the standard zero-shot cross-lingual transfer \cite{macavaney2020teaching,huang-etal-2021-improving-zero}, other cross-lingual transfer approaches, commonly applied in other NLP tasks, such as training data translation \cite{shi-etal-2020-cross}, or leveraging external word-level alignments \cite{huang2021mixed}, as well as distant supervision \cite{Yuetal2021distantsupervision} have been explored as means to improve the cross-lingual transfer of neural rankers in IR.  
While translation-based approaches are competitive for high-resource languages, they may not be as effective for low-resource languages for which reliable MT models are missing; also, translation-based cross-lingual transfer has been shown to suffer from unwanted artifacts, such as ``translationese'' \cite{zhao-etal-2020-limitations,vanmassenhove2021machine}. 

\vspace{1.5mm}
\noindent \textbf{Contributions.} Even if one would have sufficient amounts of labelled data in target languages, training language- or language-pair specific neural rerankers for all languages and language pairs would be prohibitively computationally expensive and unsustainable \cite{strubell2019energy}. In this work we additionally remedy for this by composing (re)rankers in a modular way that enables more sustainable cross-lingual transfer. Concretely, we introduce neural (re)ranking models for cross-lingual and multilingual document retrieval based on MMTs that enable much more parameter efficient fine-tuning and more effective cross-lingual transfer for relevance prediction. Our (re)rankers are based on two styles of modular components: \textbf{1)} \textit{Adapters} \cite{rebuffi2017learning,houlsby2019parameter,pfeiffer2020madx} and \textbf{2)} \textit{Sparse Fine-Tuning Masks} (SFTMs) \cite{ansell2021composable}. 
When integrated into the architecture of a pretrained MMT, both allow for (1) the pretrained multilingual knowledge to be fully preserved, alleviating the negative interference and forgetting effects, and (2) offer additional language-specific model capacity which is used to improve the MMTs' representations for target languages, thus remedying for the curse of multilinguality. 

We provide an extensive evaluation of both approaches in (i) zero-shot transfer for monolingual retrieval and (ii) CLIR, on two established benchmarks \cite{braschler2003clef,hc4}. As an additional contribution, we expand the CLEF dataset \cite{braschler2003clef} with three query languages from the Turkic family (Turkish, Kyrgyz, and Uyghur, the latter two being low-resource languages), typologically and etymologically distant from the Indo-European languages.\footnote{In this manner, our work addresses the calls for more linguistic diversity in NLP and IR research \cite{bender2011achieving,joshi2020state,ponti2020xcopa,ruder2021xtremer}.} Our results show that our modular neural (re)rankers are not only faster to train, but also outperform standard zero-shot transfer based on full MMT fine-tuning, and especially so in retrieval tasks that involve linguistically distant and low-resource languages. Moreover, our adapter- and SFTM-based rerankers generally outperform a strong preranker that utilizes state-of-the-art machine translation.





\section{Methodology}
\label{s:methodology}

We first introduce the general multi-stage ranking (i.e., preranking-reranking) framework, commonly used in information retrieval tasks, within which our work is embedded. We then introduce adapters and sparse fine-tuning masks (SFTMs), and present how to leverage them as crucial vehicles of the parameter-efficient cross-lingual transfer of the reranking component. 

\subsection{Multi-Stage Ranking}
\label{ss:multistage}

\begin{figure}
    \centering
    \includegraphics[width=0.48\textwidth]{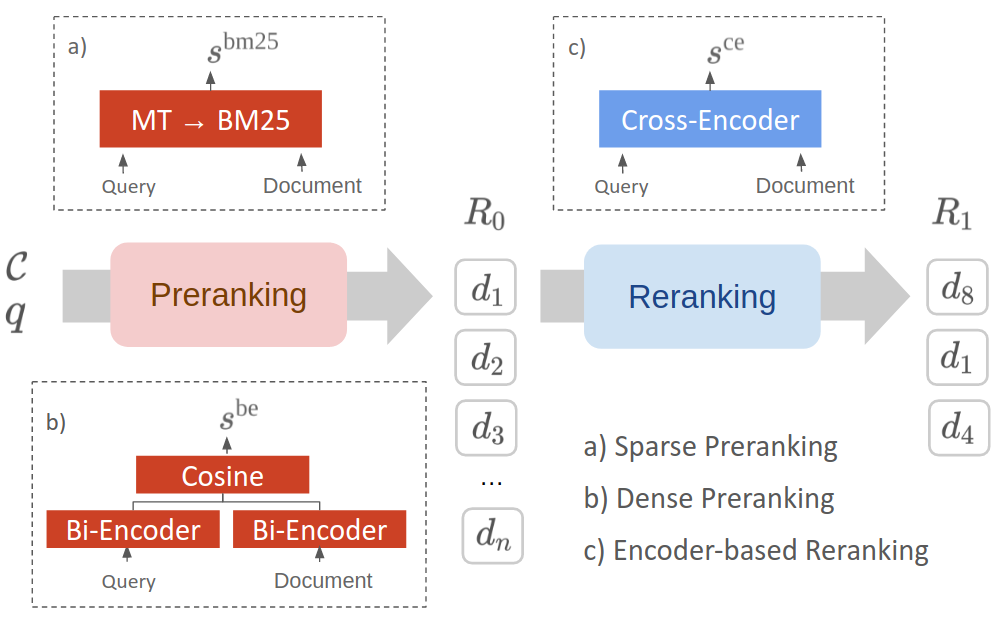}
    \caption{Overview of the multi-stage ranking approach to ad-hoc retrieval. \textbf{Stage 1 - Preranking:} We rank the document collection $\mathcal{C}$ by (a) running \textit{sparse} BM25 retrieval on translated queries, or (b) according to the cosine similarity between \textit{dense} query and document representations yielding an initial ranking $R_0$. \textbf{Stage 2 - Reranking:} We refine $R_0$ by reranking the top-k documents according to relevance scores predicted by a Cross-Encoder, yielding the refined ranking $R_1$. 
    }
    \vspace{-1mm}
    \label{fig:overview}
    \vspace{-1.5mm}
\end{figure}

Pretrained Transformers like BERT \cite{devlin-etal-2019-bert} are often used as \textit{Cross-Encoder (CE)} scoring models: the Transformer encodes a query-document concatenation fed as input to the model, and the encoding is then fed to a dense layer that predicts the relevance score \cite{macavaney2020teaching,jiang2020cross,nogueira2019multi}. 
Computing scores for all query-documents pairs with Cross-Encoders is too slow for practical IR applications: they are thus primarily used as rerankers in a multi-stage ranking approach \cite{macavaney2020teaching,Geigle:2021arxiv}. In this work we adopt this paradigm for cross-lingual ad-hoc retrieval: Figure~\ref{fig:overview} illustrates its workflow.\footnote{Alternative approaches that leverage pretrained encoders for IR include late interaction models \cite{khattab2020colbert,gao2021coil,nair2022transfer,santhanam-etal-2022-colbertv2}, embedding-based retrieval \cite{hofstaetter2021bertdot,litschko2021encoderclir}, and augmentation \cite{nogueira2019document,nogueira2019doc2query}.}  

\textit{Preranking}, based on a fast and efficient ranking method, is applied to every document from the document collection in order to provide a good initial ranking, targeting high recall.  Let $q^{l1}$ be a query in language $l_1$ and $\mathcal{C}^{l2}=\{d_i\}^n_{i=1}$ be a document collection containing $n$ documents in language $l_2$. Associating and ranking documents w.r.t. relevance scores $s_i$ we obtain an initial ranking  
\begin{align}
R_0 = [(d_1, s_1), (d_2, s_2) \dots (d_n, s_n)],
\end{align}
where $s_1 > s_2 > \dots s_n$. We transfer our rerankers based on MMTs -- and trained on English relevance judgments -- to (i) CLIR tasks as well as to (ii) monolingual IR tasks in target languages. The latter task, termed MoIR, is effectively zero-shot cross-lingual transfer for monolingual retrieval. In MoIR, we opt for a lexical preranker and score documents with $s^{\text{bm25}}=\text{BM25}(q,d)$.\footnote{We used the \texttt{pyserini} implementation of BM25 \cite{lin2021pyserini} with the suggested (i.e., default) parameter configuration.} In CLIR we follow the widely used approach of machine translating the query \cite{bonifacio2021mmarco,hc4}: this process effectively translates CLIR into a noisy variant of MoIR. In addition, we experiment with a representation-based approach based on pretrained multilingual \textit{Bi-Encoders} (\textit{BE}): here, we embed the query and documents independently, and then use the cosine similarity between their embeddings $s^{be}=cos(\mathit{BE}(q), \mathit{BE}(d))$. In the preranking stage, unlike later in reranking, we use the encoders merely as general-purpose text encoders, without any additional retrieval-specific training. 

\textit{Reranking:} This stage refines the initial ranking obtained via preranking. It relies on a $\mathit{CE}$ model which captures fine-grained (but more costly to model and run) semantic interactions between queries and documents. The ranking is then:
\begin{align}
R_1 = [(d_1, s^{ce}_1), (d_2, s^{ce}_2) \dots (d_k, s^{ce}_k)]
\end{align}
To this end, we rely on multilingual CEs to compute the binary relevance score $s^{ce}$ on the concatenation of query and document pairs: $s^{ce}=\mathit{CE}(\texttt{[CLS]} q \texttt{[SEP]} d_i \texttt{[SEP]})$. We adopt a common practice \cite{macavaney2019cedr,craswell2020overview,naseri-2021-ceqe} of reranking the top $k = 100$ pre-ranked documents, yielding the final ranking $R_1$. 
Finally, it is also possible to ensemble the preranker's and reranker's ranked lists via simple rank averaging. In our experiments (\ref{s:results}), we evaluate such preranking-reranking ensembles as well and show that such interpolations often bring additional performance gains.

\begin{figure*}
    \centering
    \includegraphics[width=\linewidth]{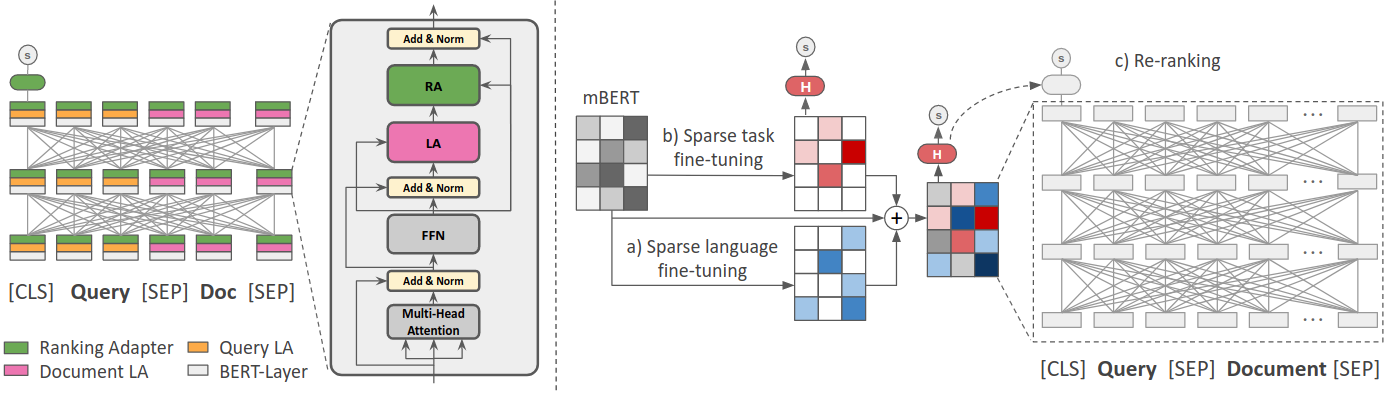}
    \caption{Overview of parameter-efficient transfer learning for neural (re)ranking. \textit{Left}: A reranker is composed by stacking a pretrained target Language Adapter (LA) and a Ranking Adapter (RA; trained with source language data) on top of the original Transformer layers of an MMT (e.g., mBERT). \textit{Right}: Sparse fine-tuning of a Ranking Mask (RM) and a Language Mask (LM) from mBERT parameters; rerankers are composed by adding the RM and LM values to the original mBERT parameters. 
    }
    \vspace{-1mm}
    \label{fig:architecture}
    \vspace{-1.5mm}
\end{figure*}


\subsection{Parameter-Efficient Cross-Lingual Ranker Transfer}
\label{ss:efficient}

In this work, we propose a modular and parameter-efficient framework that allows faster training and more effective cross-lingual transfer of neural rerankers, that enhances both CLIR and MoIR. 
We first learn language-specific Adapters (LAs) or Sparse Fine-Tuning Masks (SFTMs) via Masked Language Modelling (MLM) on unannotated monolingual corpora of respective languages, while keeping the original MMT parameters intact. We then train Ranking Adapters (or Ranking SFTMs) using source-language data on top of the source-language LAs (language SFTMs), while keeping all other parameters frozen. At inference time, for a given IR (MoIR or CLIR) task, we compose our reranker by placing the Ranking Adapters (Ranking SFTMs) on top of the LAs (language SFTMs) of the query and/or document languages of that concrete retrieval task. The modular framework is illustrated in Figure \ref{fig:architecture}.

\vspace{1.4mm} 
\noindent \textbf{Adapters.} We train Ranking Adapters (RA) and Language Adapters (LA) based on the architecture of \citet{pfeiffer2020madx}. In the Transformer architecture, each layer $l$ consists of a multi-head attention block (i.e., sub-layer) and a feed-forward network (FFN), both followed by a residual connection and layer normalization. We denote the residual connection (output of FFN) with $\mathbf{r}_l$ and the hidden state after the layer norm with $\mathbf{h}_l$. 
\begin{align}
    & \text{LA}(\mathbf{h}_l, \mathbf{r}_l) = \mathbf{U}_l (\psi (\mathbf{D}_l (\mathbf{h}_l)) + \mathbf{r}_l \\
    & \text{RA}(\mathbf{h}_l, \mathbf{r}_l) = \mathbf{U}_l (\psi (\mathbf{D}_l (\text{LA}_l))) + \mathbf{r}_l
\end{align}

\noindent Adapters are parameterized by the down-projection matrix  $\mathbf{D} \in \mathbb{R}^{h\times d}$ and the up-projection matrix $\mathbf{U} \in \mathbb{R}^{d\times h}$, where $h$ and $d$ denote the hidden size of the Transformer and the bottleneck dimension of the adapter, respectively. The ratio between $h$ and $d$ is also called the \textit{reduction factor}, and corresponds to the level of parameter compression (i.e., how many times fewer parameters are updated if we train adapters instead of updating all Transformer parameters). The forward pass of a Language Adapter consists of a down-projection of $h_l$, a non-linear activation function $\psi(\cdot)$ and an up-projection. Ranking Adapters are stacked on top of LAs and process their output. Both adapters have residual connections to the output of the FFN.\footnote{To alleviate the mismatch between the multilingual vocabulary of the MMT and the target language vocabulary, \citet{pfeiffer2020madx} also additionally place invertible adapters $\text{INV}$ on top of the embedding layer along with their inverses $\text{INV}^{-1}$ placed before the output layer. In our experiments we adopt this variant; for more details we refer the reader to the work of \citet{pfeiffer2020madx}.} We train LAs using the standard MLM objective \cite{devlin-etal-2019-bert}, whereas we train RAs together with the dense scoring layer by means of minimizing the standard binary cross-entropy loss. 

In CLIR setups, queries and documents are in different languages. It is thus, in principle, possible to stack the $\text{RA}$ on top of (i) the query language adapter $\text{LA}^{\text{Q}}$, (ii) document language adapter $\text{LA}^{\text{D}}$, or by using (iii) \textit{split adapters} $\text{LA}^{\text{S}}$: here, we encode query tokens up to the separator token (\texttt{[SEP]}) using the LA of the query language and the document tokens (after \texttt{[SEP]}) with the LA of language of the document collection (cf.~Fig. \ref{fig:architecture}). 

\vspace{1.4mm}
\noindent \textbf{Sparse Fine-Tuning Masks.} Like adapters, SFTMs \cite{ansell2021composable} aim to decouple task knowledge from language knowledge, but instead of introducing additional parameters, the idea is to directly update only small subsets of MMT's original parameters. 
Sparse Fine-Tuning (SFT) consists of two phases. In \textit{Phase 1} we fine-tune all mBERT's parameters $\theta^{(0)}$, resulting in updated parameter values $\theta^{(1)}$. We then select the top $K$ parameters with the largest value change, i.e., those with the largest values $|\theta_i^{(0)} - \theta_i^{(1)}|$. We then construct a binary mask: the selected $K$ parameters remain trainable, whereas all other parameters are frozen. In \textit{Phase 2} all parameters are reset to $\theta^{(0)}$ and training restarts, but this time only the selected parameters of the mask are updated, yielding $\theta^{(2)}$. The final update (i.e., the SFTM) is then obtained as the difference vector $\text{M}=\theta^{(2)}-\theta^{(0)}$. As is the case with Language Adapters, we obtain the Language Masks (LM) by means of (additional) MLM training on language-specific corpora; whereas the Ranking Mask (i.e., the mask for the ranking task, $\text{RM}$) is learned via binary cross-entropy objective on source-language (English) relevance judgments. 
At inference, the reranker is composed as $\theta^{(0)} + \text{RM} + \text{LM}$ (cf., Figure \ref{fig:architecture}). In our CLIR settings (§\ref{s:experimental}), we explore using (i) the query language mask ($\text{LM}^{\text{Q}}$), (ii) document language mask ($\text{LM}^{\text{D}}$) or (iii) the combination of both masks ($\text{LM}^{\text{B}} = \text{LM}^{\text{Q}} + \text{LM}^{\text{D}}$). Note that SFTMs represent a more computationally efficient solutions at inference time: unlike adapters, they do not extend (i.e., deepen) the Transformer architecture.

\section{Experimental Setup}
\label{s:experimental}

\noindent\textbf{Adapter and SFTM Training.} We train adapters following the recommendations from \citet{pfeiffer2020madx}. Unless noted otherwise, we train $\text{LAs}$ with the reduction factor of $2$ (i.e., $h/d = 2$) on Wikipedias of respective languages, for $250$K steps with batch size $64$ and learning rate of $1$e-$4$. For RAs we experimented with the different reduction factors: 1, 2, 4, 8, 16, 32 (cf. §\ref{s:results}). Following \citet{ansell2021composable}, for fair comparisons between adapters and SFTMs, we set the mask size $K$ for SFTMs to the same number of parameters that adapters with a certain reduction factor have.\footnote{Leading to the number of trainable parameters (sparsity) of 14M (8.5\%), 7.1M (4.2\%), 3.6M (2.1\%), 1.8M (1.1\%), 894K (0.52\%) and 452K (0.27\%) respectively.}

\vspace{1.5mm}
\noindent\textbf{Reranking Training.}
We train mBERT-based\footnote{Pretrained \texttt{bert-base-multilingual-uncased} weights from the HuggingFace Transformers library \cite{hf_library} are used.} rerankers on MS-MARCO \cite{craswell2021ms}, with a linear warm-up over the first 5K updates, in batches of $32$ instances with a maximum sequence length of $512$, and using a learning rate of $2$e-$5$. We evaluate the model on the validation data every 25K updates and choose the checkpoint with the best validation performance.

\vspace{1.5mm}
\noindent\textbf{Evaluation Data.} 
We evaluate the models on the standard CLEF-2003 benchmark \cite{braschler2003clef}\footnote{\url{http://catalog.elra.info/en-us/repository/browse/ELRA-E0008/}} as well as on the recently introduced HC4 benchmark \cite{hc4}. With CLEF, we use monolingual test collections in EN, DE, IT, RU, and FI for MoIR, and experiment with the following cross-lingual directions: EN-\{FI, DE, IT, RU\}, DE-\{FI, IT, RU\}, FI-\{IT, RU\}. Each experimental run covers 60 queries, whereas the document collection sizes are as follows: RU -- 17K, FI -- 55K, IT -- 158K, and DE -- 295K. 

We additionally evaluate the models in CLIR tasks with CLEF queries posed in lower-resource languages. To this end, (i) we leverage Swahili (SW) and Somali (SO) queries \cite{Bonab2019swahiliclef}, where the queries were obtained via manual translation of English queries; (ii) we create another set of translated CLEF queries in three languages: Turkish (TR), Kyrgyz (KG), and Uyghur (UG). The new set covers one high-resource and two low-resource languages and is intended to facilitate and diversify evaluation of CLIR with low-resource languages in future work. The queries were constructed via the standard post-editing procedure borrowed from other data collection tasks \cite{glavavs2020xhate,hung-etal-2022-multi2woz}: we obtained initial query translations via Google Translate, which were then post-edited by native speakers.

HC4 comprises queries \textit{and} document collections in three languages: Persian (FA), Russian (RU) and Chinese (ZH). Compared to CLEF, HC4 collections are considerably larger, spanning 646K, 486K and 4.72M documents per each respective language, associated with 50 test queries in each language. We use \textit{title} and \textit{description} fields as queries following \citet{hc4}. HC4 is used in MoIR experiments.

\setlength{\tabcolsep}{2.1pt}
\begin{table*}[ht!]
{
\small
\begin{tabularx}{1\linewidth}{l ccccc cccc ccc cc cc}
\toprule
Model & {\scriptsize TR-EN} & {\scriptsize TR-IT} & {\scriptsize TR-DE} & {\scriptsize TR-FI} & {\scriptsize TR-RU} & {\scriptsize EN-FI} & {\scriptsize EN-IT} & {\scriptsize EN-RU} & {\scriptsize EN-DE} & {\scriptsize DE-FI} & {\scriptsize DE-IT} & {\scriptsize DE-RU} & {\scriptsize FI-IT} & {\scriptsize FI-RU} & {\scriptsize \textbf{AVG}} & {\scriptsize \textbf{ENS}} \\ \midrule
$\text{DISTIL}_{\text{DmBERT}}$ (PR) & .183 & .251 & .190 & .252 & .260 & .294 & .290 & .313 & .247 & .300 & .267 & .284 & .221 & .302 & .261 & -\\ 
MonoBERT & .235 & .197 & .208 & .285 & .217 & .339 & .315 & .248 & .295 & .329 & .270 & .246 & .197 & .174 & .254 & .274 \\  \midrule
$+\text{RA}$ $+\text{LA}^{\text{S}}$ & .269 & \textbf{.253} & \textbf{.252}* & \textbf{.362} & .186 & .363 & .352 & .197 & .317* & .329 & .300 & .223 & \textbf{.266} & .207 & .277 & .287 \\
$+\text{RA}$ $+\text{LA}^{\text{D}}$ & .252 & .234 & .222 & .267 & \textbf{.267} & .366* & \textbf{.366}* & \textbf{.248} & .314* & .350 & .302 & \textbf{.315} & .220 & \textbf{.234} & \textbf{.283} & \textbf{.298} \\
$+\text{RA}$ $+\text{LA}^{\text{Q}}$ & \textbf{.270} & .243 & .242 & .293 & .191 & .370 & .355 & .189 & .318 & .325 & .279 & .223 & .247 & .182 & .266 & .285 \\ \noalign{\vskip 0.2ex}\cdashline{1-17}[.4pt/1pt]\noalign{\vskip 0.5ex}
$+\text{RM}$ $+\text{LM}^{\text{B}}$ & .229 & .228 & .197 & .244* & .168 & .299 & .344 & .181* & .303 & .309 & .302 & .191* & .206 & .108* & .236 & .269 \\
$+\text{RM}$ $+\text{LM}^{\text{D}}$ & .231 & .226 & .229 & .317 & .149* & \textbf{.394*} & .359 & .173* & .320* & .376 & .304 & .187 & .239 & .166* & .262 & .279 \\
$+\text{RM}$ $+\text{LM}^{\text{Q}}$ & .239 & .252 & .232 & .316 & .162* & .359 & .349 & .191 & .310* & \textbf{.391} & \textbf{.323*} & .195 & .255* & .160 & .267 & .280 \\ \bottomrule
\end{tabularx}
}
\caption{CLIR results (Mean Average Precision, MAP) with $\text{DIST}_{\text{DmBERT}}$ as Stage 1 preranker. \textbf{Bold}: Best neural retrieval model for each language pair. *: significance tested against MonoBERT at $p \leq 0.05$, computed via paired two-tailed t-test. Ranking and Language Adapters have a reduction factor of 16 and 2 (see \S\ref{s:methodology}), respectively. Ranking and Language Masks both correspond to a reduction factor of 2 (see \S\ref{s:experimental}). We report average results (AVG), and also averaged ensemble (ENS) results where we combine ranking lists from Stage 1 and Stage 2 rankings; see \S\ref{ss:multistage}. Superscripts over LAs and LMs denote query language (Q), document language (D), split adapters (S) for LAs, and `(B)oth masks' for LMs (see \S\ref{ss:efficient}).}  
\label{tbl:clir}
\end{table*}

\setlength{\tabcolsep}{2.5pt}
\begin{table*}[ht!]
{
\small
\begin{tabularx}{1\linewidth}{l ccccc cccc ccc cc cc}
\toprule
Model & {\scriptsize TR-EN} & {\scriptsize TR-IT} & {\scriptsize TR-DE} & {\scriptsize TR-FI} & {\scriptsize TR-RU} & {\scriptsize EN-FI} & {\scriptsize EN-IT} & {\scriptsize EN-RU} & {\scriptsize EN-DE} & {\scriptsize DE-FI} & {\scriptsize DE-IT} & {\scriptsize DE-RU} & {\scriptsize FI-IT} & {\scriptsize FI-RU} & {\scriptsize \textbf{AVG}} & {\scriptsize \textbf{ENS}} \\ \midrule
NMT+BM25 (PR) & .392 & .353 & .308 & .307 & .227 & .378 & .446 & .285 & .355 & .367 & .385 & .272 & .364 & .271 & .336 & -\\
MonoBERT & .415 & .375 & .339 & .345 & .307 & .386 & .411 & .351 & .371 & .409 & .380 & .322 & .367 & .340 & .366 & .360 \\ \midrule
 +RA +LA & \textbf{.448} & .408* & .353 & .371 & .327 & .388 & \textbf{.435} & \textbf{.367} & .385 & .413 & .405 & .348 & .381 & \textbf{.365} & .385 & \textbf{.374} \\
 +RM +LM & .447 & \textbf{.414} & \textbf{.356} & \textbf{.386} & \textbf{.336} & \textbf{.413} & .429 & .345 & \textbf{.390} & \textbf{.468} & \textbf{.407} & \textbf{.363} & \textbf{.395} & .364 & \textbf{.394} & .371\\ \bottomrule
\end{tabularx}
}
\caption{CLIR results (Mean Average Precision, MAP) with NMT+BM25 as Stage 1 preranker. For modular rerankers, we report the numbers with the best-performing configurations from CLEF experiments: $+\text{RA}$ $+\text{LA}^{\text{D}}$ and $+\text{RM}$ $+\text{LM}^{\text{Q}}$; see also the caption of Table~\ref{tbl:clir}.}  
\label{tbl:ltir}
\vspace{-0.5em}
\end{table*}

\vspace{1.5mm}
\noindent\textbf{Baseline Models.} The primary baseline for our adapter- and SFTM-based transfer is the standard and well established method for zero-shot transfer of English-trained rerankers \cite{macavaney2020teaching}, termed \textit{MonoBERT}. This is the reranking Cross-Encoder where we allow for full-tuning of the underlying monolingual or multilingual BERT model on MS-MARCO. For CLIR experiments, we opt for $\text{DISTIL}_{\text{DmBERT}}$ as our Bi-Encoder preranker (PR), as it showed strong performance in our recent comparative empirical study \cite{litschko2021encoderclir}. In brief, $\text{DISTIL}_{\text{DmBERT}}$ is trained via knowledge distillation where sentence-similarity features are distilled from a monolingual English teacher, specialized for semantic encoding of sentences, into a multilingual student model; see \cite{reimers2020distil} for further details. 

Finally, also for CLIR, we couple a state-of-the-art NMT system of \newcite{fan2020beyond} (FAIR-MT), which we use to translate queries to the document collection language, with the BM25 ranker in the target language. For Kyrgyz and Uyghur, we use another NMT model, provided by the Turkic Interlingua (TIL) community\footnote{\url{https://turkic-interlingua.org}} \cite{mirzakhalov2021large}, because we failed to obtain meaningful \{KG, UG\}$\rightarrow$ $l_2$ translations with FAIR-MT. 




\section{Results and Discussion}
\label{s:results}

%
%
%
%
%

\setlength{\tabcolsep}{7.5pt}
\begin{table*}
\centering
\small
\def\arraystretch{0.85}
\begin{tabular}{l   c c c c c c    c c c}
\toprule
& \multicolumn{4}{c}{CLEF 2003} & \multicolumn{3}{c}{HC4}  \\ \cmidrule(lr){2-5}\cmidrule(lr){6-8}
Model & SW--EN & SO--EN & KG--EN & UG--EN & EN--FA & EN--ZH & EN--RU & \textbf{AVG} & \textbf{ENS} \\ \midrule
NMT+BM25 (PR) & .325 & .157 & .228 & .091 & .183 & .113 & .186 & .183 & - \\
MonoBERT & .362 & .158 & .255 & .157 & .246 & .172 & .218  & .224 & .216 \\ \noalign{\vskip 0.2ex}\cdashline{1-10}[.4pt/1pt]\noalign{\vskip 0.5ex}
$+\text{RA} +\text{LA}^{\text{D}}$ & \textbf{.407} & \textbf{.166} & .305 & .155 & .259 & .189 & .234 & .245 & \textbf{.228}  \\
$+\text{RM} +\text{LM}^{\text{D}}$ & .389 & .161 & \textbf{.311} & \textbf{.165} & \textbf{.267} & \textbf{.196} & \textbf{.241} & \textbf{.247} & .225 \\
\bottomrule 
\end{tabular} 
\caption{CLIR results on extended CLEF pairs with low-resource query languages (Swahili, Somali, Kyrgyz, and Uyghur) and three language pairs from the HC4 benchmark.}
\label{tbl:lowres}
\vspace{-1mm}
\end{table*}

\noindent \textbf{Cross-Lingual Retrieval (CLIR).} 
Tables~\ref{tbl:clir} and \ref{tbl:ltir} show the CLIR results, for fourteen language pairs from the augmented CLEF 2003 benchmark\footnote{We add TR-* pairs to the evaluation, enabled by our EN$\rightarrow$TR translations of the queries.} using $\text{DISTIL}_{\text{DmBERT}}$ and NMT+BM25 as Stage 1 prerankers, respectively. 
With $\text{DISTIL}_{\text{DmBERT}}$ as the preranker (Table \ref{tbl:clir}), Adapter- and SFTM-based rerankers consistently improve the initial preranking results, with gains of up to 2.7 MAP points, and EN-RU as the only exception.
Importantly, compared to full fine-tuning (MonoBERT), our modular reranking variants bring gains between 1 and 4 MAP points on average, across all language pairs.   
Interestingly, the best adapter configuration ($\text{RA}$ $+\text{LA}^{\text{D}}$), where at inference we stack the RA on top of the LA of the document collection language) outperforms the best SFTM-based reranker ($\text{RM}$ $+\text{LM}^{\text{Q}}$ and $\text{RM}$ $+\text{LM}^{\text{D}}$) by 1.6 MAP points. 
Somewhat surprisingly, adapting only to the language of the document collection ($\text{LA}^{\text{D}}$; $\text{LM}^{\text{D}}$) yields better performance than adapting to both the query and collection language of the target task ($\text{LA}^{\text{S}}$; $\text{LM}^{\text{B}}$). 

The language pairs in Tables \ref{tbl:clir} and \ref{tbl:ltir} consist of high-resource languages for which large parallel corpora and, consequently, reliable NMT models exist. However, even when starting from a more competitive MT-based preranker (NMT+BM25; Table \ref{tbl:ltir}), our modular cross-lingual transfer of the reranker yields performance gains. In fact, with this stronger preranker, the gains from modular reranking are even more pronounced: +5/+6 MAP points for Adapters and SFTMs, respectively, compared to preranker and +2/+3 MAP points, respectively, compared to MonoBERT. This could explain why interpolating between the preranking and reranking (ENS, last column) yields further gains with $\text{DISTIL}_{\text{DmBERT}}$ as the preranker (Table \ref{tbl:clir}), but not when we prerank with NMT+BM25 (Table \ref{tbl:ltir}).

Table~\ref{tbl:lowres} shows CLIR results for (a) language pairs from extended CLEF with queries written in low-resource languages -- Swahili and Somali queries created by \newcite{Bonab2019swahiliclef}, as well as Kyrgyz and Uyghur queries that we created; and (b) three cross-lingual pairs of arguably distant languages (EN-\{Farsi, Chinese, Russian\}) from the HC4 benchmark. The gains that our SFTM- and Adapter-based modular rerankers bring for language pairs involving low-resource languages, over the MT-based preranker and the full fine-tuning (MonoBERT), are generally more substantial than those for high-resource language pairs: e.g., +8 and +4 MAP points w.r.t. NMT+BM25 and MonoBERT, respectively for SW-EN, and +8 and +5 points for KG-EN. The gains are similarly prominent for more distant language pairs from the HC4 dataset (+8 MAP points over the NMT+BM25 preranker for EN-FA and EN-ZH). With such prominent gains of the modular reranking over the preranker, it is no surprise that averaging the preranking and reranking document ranks (ENS) reduces the performance of the reranker.       
%
%
%
%
We believe that these results in particular emphasize the effectiveness of modular cross-lingual transfer that allows to increase the capacity of MMTs for individual languages, by means of LMs or LAs. The representations of low-resource languages, for which MMTs have seen little data in pretraining, particularly suffer from the curse of multilinguality \cite{conneau2020unsupervised,lauscher2020zero} -- this is why particularly prominent gains are achieved for those languages when we increase the MMTs capacity for their representation via LMs/LAs.      




\setlength{\tabcolsep}{11pt}
\begin{table*}[ht!]
\centering
\small
\begin{tabular}{l   c c c c c c    c c c   c}
\toprule
& \multicolumn{5}{c}{CLEF 2003} & \multicolumn{3}{c}{HC4} & \\ \cmidrule(lr){2-6}\cmidrule(lr){7-9}
Model & EN & FI & DE & IT & RU & FA & ZH & RU & \textbf{AVG} & \textbf{ENS} \\ \midrule
BM25 (PR) & .480 & .505 & .434 & .494 & .361 & .279 & .196 & .228 & .372 & - \\
MonoBERT & .464 & .528 & .444 & .463 & .363 & .356 & .283 & .245 & .398 & .402 \\ \midrule
$+\text{RA} +\text{LA}$ & .512 & .537 & .457 & .495 & \textbf{.389} & .372 & .284 & .261 & .413 & .410 \\
$+\text{RM} +\text{LM}$ & \textbf{.515} & \textbf{.564} & \textbf{.459} & \textbf{.502} & .379 & \textbf{.398} & \textbf{.307} & \textbf{.264} & \textbf{.423} & \textbf{.417} \\
\bottomrule 
\end{tabular} 
\caption{Results of zero-shot cross-lingual transfer for monolingual retrieval (MoIR) on CLEF 2003 and HC4 datasets. Results with reduction factors of 16 and 2 for Adapters and SFTMs, respectively.}
\label{tab:moir}
\vspace{-1mm}
\end{table*}

\vspace{1.4mm}
\noindent \textbf{Cross-Lingual Transfer for MoIR.} 
Table~\ref{tab:moir} displays the results of monolingual retrieval with our best-performing modular rerankers for EN (as the source language) and four target languages (DE, IT, FI, RU).\footnote{Note that in MoIR, the actual reranking is always monolingual (albeit in the target language). Both queries and documents are thus encoded with the same target language LA/LM.}
%
%
%
%
Unlike the fully fine-tuned reranker (MonoBERT), our modular Adapter- and SFTM-based rerankers improve the initial rankings produced by BM25. These results strengthen the finding that our modular rerankers are not just more parameter-efficient (i.e., faster to train), but also lead to better cross-lingual transfer due to decoupling of language- and ranking-specific knowledge. 
In MoIR tasks the SFTM-based transfer outperforms its Adapter-based counterpart, same as in the case of CLIR with NMT+BM25 preranking (Table \ref{tbl:clir}). Also as in the case of the latter CLIR results (Tables \ref{tbl:clir} and \ref{tbl:lowres}), interpolating between preranking and reranking results does not bring any gains. 

It is worth noting that all MoIR scores are substantially higher than CLIR results from Tables~\ref{tbl:clir} and \ref{tbl:ltir}. This is expected and reflects the fact that matching representations within a language -- where models can still rely on exact lexical matches between queries and documents -- is easier than aligning text representations across languages.   

\setlength{\tabcolsep}{2.5pt}
\begin{table}
{
\small
\begin{tabular}{l c c c c c c}
\toprule
Layer & CLIR & MoIR & AVG & Latency & $\Delta$~Speed-Up & $\Delta$~MAP \\ \midrule
None & .282 & .418 & .331 & 34.6 ms & -	& -\\ \noalign{\vskip 0.2ex}\cdashline{1-7}[.4pt/1pt]\noalign{\vskip 0.5ex}
1-2 & .295 & .412 & .337 & 33.7 ms & $+$2.6\% & $+$.006 \\
1-4 & .269 & .395 & .314 & 32.8 ms & $+$5.0\% & $-$.017\\ 
1-6 & .229 & .375 & .281 & 31.9 ms & $+$7.7\% & $-$.050 \\ 
1-8 & .134 & .284 & .187 & 31.0 ms & $+$10.4\% & $-$.143 \\ 
1-10 & .086 & .210 & .130 & 30.0 ms & $+$12.9\% & $-$.200 \\ 
1-12 & .086 & .208 & .129 & 29.5 ms & $+$14.2\% & $-$.201 \\ 
\bottomrule 
\end{tabular} 
}
\caption{Trade-off between efficiency and effectiveness when dropping adapters in $+\text{RA}+\text{LA}^{\text{D}}$. Average over all CLIR/MoIR setups and all reduction factors.}
\label{tab:adapterdrop}
\end{table}

\begin{figure}[ht!]
\centering
\includegraphics[scale=.45]{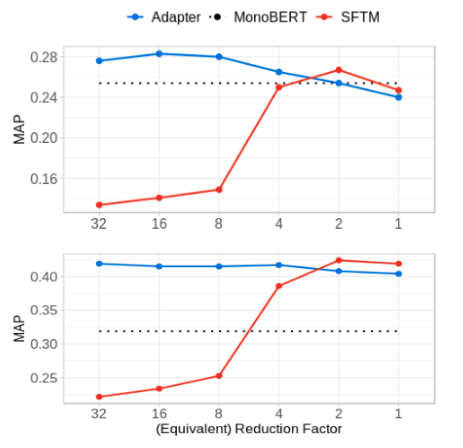}
\caption{Retrieval performance at different parameter reduction factors; average MAP performance for CLIR (top) and MoIR (bottom). 
}
\label{fig:reductionfactor}
\vspace{-0.5em}
\end{figure}








\vspace{1.5mm}
\noindent \textbf{Effectiveness vs Efficiency.} 
Adapters increase query latency because they deepen the Transformer.  \citet{ruckle-etal-2021-adapterdrop} show that one can drop adapters from lower layers with small-to-negligible effect on performance. Table~\ref{tab:adapterdrop} shows the results of a similar analysis, where we drop the adapters from the first $N$ layers at inference. Dropping adapters from only the first two layers (row 1-2) only slightly decreases the MoIR performance whereas it even slightly increases the CLIR results. Dropping adapters from more layers, however, substantially reduces the retrieval performance: e.g., removing adapters from the first 10 layers reduces CLIR performance by almost 20 MAP points, while reducing the query latency by only 13\%. While Adapters and SFTMs yield comparable performance in our experiments, these observations favor SFTMs: for the \textit{same query latency},\footnote{The query latency of an SFTM-based reranker is the same as that of MonoBERT as SFTMs do not increase the number of layers (nor parameters within layers) of the MMT.} SFTMs will yield better performance. 

\vspace{1.5mm}
\noindent \textbf{Parameter Efficiency.} We also investigate the relation between various levels of parameter efficiency and retrieval performance. Figure~\ref{fig:reductionfactor} shows the performance of our modular rerankers for different parameter reduction factors. SFTMs exhibit stronger performance with smaller reduction factors (2 and 4), i.e., when we update a larger percentage of mBERT's original parameters. SFTMs shift the pretrained values of mBERT's parameters: this constrains the range of values that individual parameters can take, requiring the modification of the larger number of parameters for injecting complex language- and ranking-specific knowledge. 
In contrast, Adapters show better performance with higher reduction factor (8, 16, 32), i.e., when we add a relatively smaller number of Adapter parameters. This could be the consequence of the ``unconstrained'' initialization of the new Adapter parameters, which allows the complementary language- and ranking-specific knowledge to be compressed into a smaller number of parameters.    
Comparing those effects between CLIR and MoIR we observe the same trends. However, MAP gains compared to MonoBERT are larger in MoIR than in CLIR. This seems intuitive as ranking adapters (masks) are able to adapt for exact matches. 


\setlength{\tabcolsep}{5pt}
\begin{table*}
{
\def\arraystretch{0.85}
\small
\begin{tabular}{m{1cm} p{4.5cm} p{4.5cm} p{4.5cm}}
\toprule
QID & English Query \textit{(original)} & NMT: Swahili $\rightarrow$ English & NMT: Somali $\rightarrow$ English \\ \midrule
151 & Wonders of \textbf{Ancient World}   Look for \textbf{information} on the \textbf{existence} and/or the discovery of remains of the seven wonders of the \textbf{ancient world}. & Search for \textbf{information} about the \textbf{existence} and/or development of the \textbf{seventh} universe of the \textbf{ancient world}. & Thus, therefore, it is necessary to bear in mind that the truth is the truth, and that the truth is the truth, and that the truth is the truth. \\ \noalign{\vskip 0.5ex}\cdashline{1-4}[.4pt/1pt]\noalign{\vskip 0.5ex}
172 & \textbf{1995} Athletics \textbf{World Records}   What \textbf{new world records} were achieved during the \textbf{1995} athletic world championships in \textbf{Gothenburg}? & What \textbf{new world records} were recorded at the \textbf{1995} \textbf{World} Horses in \textbf{Gothenburg}? & The \textbf{1995} \textbf{World} Trade Organization (WTO) announced that a \textbf{new} international trade agreement has led to a global trade agreement in \textbf{Gothenburg}. \\ \noalign{\vskip 0.5ex}\cdashline{1-4}[.4pt/1pt]\noalign{\vskip 0.5ex}
187 & \textbf{Nuclear} Transport in \textbf{Germany} Find reports on the protests against the transportation of \textbf{radioactive} waste with \textbf{Castor} \textbf{containers} in \textbf{Germany}. & \textbf{Nuclear} Delivery in \textbf{Germany} A report on the anti-trafficking of \textbf{radioactive} pollutants and \textbf{Castor} \textbf{containers} in \textbf{Germany}. & The Nugleerka department of Jarmalka Hel has been prepared for the development of the Nugleerka department of \textbf{Castor} district in Jarmalka. \\ \noalign{\vskip 0.5ex}\cdashline{1-4}[.4pt/1pt]\noalign{\vskip 0.5ex}
200 & Flooding in Holland and \textbf{Germany}   Find statistics on flood disasters in Holland and \textbf{Germany} in \textbf{1995}. & The floods in the Netherlands and \textbf{Germany} have recorded the floods in the Netherlands and \textbf{Germany} in \textbf{1995}. & The Netherlands Federation and the United Nations have agreed with the Netherlands Federation and the Netherlands Federation in \textbf{1995}. \\ 
\bottomrule 
\end{tabular} 
}
\caption{Comparison between original CLEF queries and translations from Swahili and Somali to English. Tokens that occur both in the original query and translations are highlighted in \textbf{bold} (ignoring case, excluding stopwords).} 
\label{tab:qualitative}
\end{table*}

\vspace{1.4mm}
\noindent \textbf{Impact of NMT on CLIR.} In the cross-lingual setup the quality of retrieved documents crucially depends on the quality of query translations when NMT is used. In Table~\ref{tab:qualitative} we show original English queries together with their respective translations from Swahili and Somali. As expected, translations from Swahili are generally of higher quality compared to Somali, which explains the big performance gap reported in Table~\ref{tbl:lowres}. In the best case the translation is semantically very close to the original query (cf.,  {\footnotesize SW$\shortrightarrow$EN; QID:172}), or it contains only slight lexical (\textit{flooding} vs. \textit{floods}) and semantic variations, e.g., near-synonyms (\textit{Holland} vs. \textit{Netherlands}). In other cases, error propagation from NMT impacts CLIR performance to different extents. Those include, e.g., missing keywords (\textit{statistics}; {\footnotesize QID:200}), topic shifts (\textit{sports} vs. \textit{business}; {\footnotesize SO$\shortrightarrow$EN, QID:172}) or queries consisting of unrelated text and repetitions (i.e., `hallucinations'; {\footnotesize SO$\shortrightarrow$EN, QID:151, QID:200}). Especially repetitions and hallucinations\footnote{This phenomenon has been reported to occur in low-resource and out-of-domain settings \citep{muller-etal-2020-domain}. We confirm this finding as we find hallucinations appearing more often in EN$\shortrightarrow$SO than in EN$\shortrightarrow$SW query translations.} are known unwanted artifacts in NMT \citep{Fu_Lam_So_Shi_2021,raunak-etal-2021-curious} and can cause retrieval models to emphasize unrelated keywords by inflating their term frequency.\footnote{Further investigation of NMT+BM25 on SO$\shortrightarrow$EN reveals that manually filtering out queries containing more than two repetitions/hallucinations leaves us with 22 remaining queries on which results improve from 0.157 to 0.280 MAP.} Lastly, in cases where source words are copied instead of translated, e.g.,  \textit{Nugleerka (Nuclear)} or \textit{Jarmalka (Germany)} in {\footnotesize QID:187}, neural retrieval models need to rely on imperfect internal alignment of word translations \cite{cao2019multilingual}. 






\section{Related Work}
\label{s:rw}

Next to Adapters and SFTMs there exist other parameter efficient transfer (PET) methods. For example, BitFit trains only bias vectors \cite{ben-zaken-etal-2022-bitfit}, LoRa trains low-rank decompositions of weight matrices in dense layers \citep{hu2022lora} and methods that learn continuous prompts \citep[\textit{inter alia}]{liu2021p,lester-etal-2021-power,li-liang-2021-prefix}. In the context of retrieval for English, concurrent work focuses on the learning-efficiency \cite{ma2022scattered} and out-of-domain generalization \cite{WLTam2022PT-Retrieval} of PET methods, whereas we investigate PET both on task- and language-level adaption for CLIR. 

\section{Conclusion}
\label{s:conclusion}


In this work, we introduced modular and parameter-efficient neural rerankers for effective cross-lingual retrieval transfer. Our models, based on Adapters and Sparse Fine-Tuning Masks, allow for decoupling of language-specific and task-specific (i.e., ranking) knowledge. We demonstrate that this leads to more effective transfer to cross-lingual IR setups as well as to better cross-lingual transfer for monolingual retrieval in target languages with no relevance judgment improving over strong prerankers based on state-of-the-art NMT. Encouragingly, we observe particularly pronounced gains for low-resource languages included in our evaluation. 
We hope that our results will encourage a broader investigation of parameter-efficient neural retrieval in monolingual and cross-lingual setups. We make our code and resources available at: \url{https://github.com/rlitschk/ModularCLIR}.

\section*{Acknowledgments}
RL and GG are supported by the EUINACTION grant from NORFACE Governance (462-19-010, GL950/2-1). IV is supported by a Huawei research donation to the University of Cambridge.

\bibliography{references}
\bibliographystyle{acl_natbib}




\end{document}